\documentclass[10pt,twocolumn,letterpaper]{article}

\usepackage[pagenumbers]{cvpr} 
\usepackage{graphicx}
\usepackage{amssymb}

\usepackage{amsmath,amsfonts,bm,bbm}

\usepackage[usenames,dvipsnames]{xcolor}
\definecolor{snsgray}{RGB}{179, 179, 179}
\definecolor{snsorange}{RGB}{252, 141, 98}
\definecolor{snsblue}{RGB}{141, 160, 203}

\definecolor{coolgrey}{RGB}{157,157,157}
\definecolor{lightgrey}{RGB}{235,238,238}
\definecolor{lightteal}{RGB}{198,211,222}
\definecolor{cyan}{RGB}{136, 204, 238}
\definecolor{teal}{RGB}{68, 170, 153}
\definecolor{sand}{RGB}{221, 204, 119}
\definecolor{rose}{RGB}{204, 102, 119}
\definecolor{red}{RGB}{250, 94, 91}
\definecolor{orange}{RGB}{255, 200, 63}
\definecolor{yellow}{RGB}{254, 239, 109}

\definecolor{darkgreen}{rgb}{0.09, 0.45, 0.27}

\usepackage{xfrac}

\def\eqref#1{equation~\ref{#1}}

\def\1{\bm{1}}

\DeclareMathAlphabet{\mathsfit}{\encodingdefault}{\sfdefault}{m}{sl}
\SetMathAlphabet{\mathsfit}{bold}{\encodingdefault}{\sfdefault}{bx}{n}

\usepackage{booktabs}
\usepackage{marvosym}
\usepackage{xcolor}
\definecolor{cvprblue}{rgb}{0.21,0.49,0.74}
\usepackage[pagebackref,breaklinks,colorlinks,citecolor=cvprblue]{hyperref}
\usepackage[capitalize]{cleveref}
\usepackage{algorithm}
\usepackage{algpseudocode}
\usepackage{algorithmicx}
\usepackage{tcolorbox}
\usepackage{listings}
\usepackage{multirow}
\definecolor{codegreen}{rgb}{0,0.6,0}
\definecolor{codegray}{rgb}{0.5,0.5,0.5}
\definecolor{codepink}{RGB}{252, 142, 172}
\definecolor{codepurple}{rgb}{0.58,0,0.82}
\definecolor{backcolour}{RGB}{245,245,245}
\lstdefinestyle{mystyle}{
    backgroundcolor=\color{backcolour},   
    commentstyle=\color{magenta},
    keywordstyle=\color{blue},
    numberstyle=\tiny\color{codegray},
    stringstyle=\color{codepurple},
    basicstyle=\fontfamily{\ttdefault}\footnotesize,
    breakatwhitespace=false,         
    breaklines=true,                 
    keepspaces=true,    
    frame=single,
    numbersep=5pt,                  
    showspaces=false,                
    showstringspaces=false,
    showtabs=false,                  
    tabsize=2,
    classoffset=1, 
    keywordstyle=\color{violet},
    classoffset=0,
}
\lstdefinelanguage{JavaScript}{
  keywords={typeof, new, true, false, catch, function, return, null, catch, switch, var, if, in, while, do, else, case, break},
  keywordstyle=\color{blue}\bfseries,
  ndkeywords={class, export, boolean, throw, implements, import, this},
  ndkeywordstyle=\color{darkgray}\bfseries,
  identifierstyle=\color{black},
  sensitive=false,
  comment=[l]{//},
  morecomment=[s]{/*}{*/},
  commentstyle=\color{purple}\ttfamily,
  stringstyle=\color{red}\ttfamily,
  morestring=[b]',
  morestring=[b]"
}
\tcbuselibrary{skins, breakable}
\definecolor{redhl}{HTML}{FF0000}
\definecolor{greenhl}{HTML}{00FF00}
\definecolor{bluehl}{HTML}{00FFFF}
\definecolor{greyhl}{HTML}{C0C0C0}
\crefname{section}{Sec.}{Secs.}
\Crefname{section}{Section}{Sections}
\Crefname{table}{Table}{Tables}
\crefname{table}{Tab.}{Tabs.}

\def\paperID{2}
\def\confName{CVPR}
\def\confYear{2024}

\begin{document}

\title{STEVE Series: Step-by-Step Construction of Agent Systems in Minecraft}

\author{Zhonghan Zhao$^{1,*}$, Wenhao Chai$^{2,*,\dagger}$, Xuan Wang$^{1}$, Ke Ma$^{1}$, Kewei Chen$^{3}$ \\
Dongxu Guo$^{3}$, Tian Ye$^{4}$, Yanting Zhang$^{3}$, Hongwei Wang$^{1}$ and Gaoang Wang$^{1,\text{\Letter}}$ \vspace{6pt} \\
$^{1}$ Zhejiang University \quad
$^{2}$ University of Washington \quad
$^{3}$ Donghua University \\
$^{4}$ Hong Kong University of Science and Technology (GZ) \\
\small\texttt{\{zhonghan.22, gaoangwang\}@intl.zju.edu.cn, wchai@uw.edu}\\
}

\twocolumn[{
    \maketitle
    \renewcommand\twocolumn[1][]{#1}
    \begin{center}
    \centering
    \includegraphics[width=1.0\textwidth]{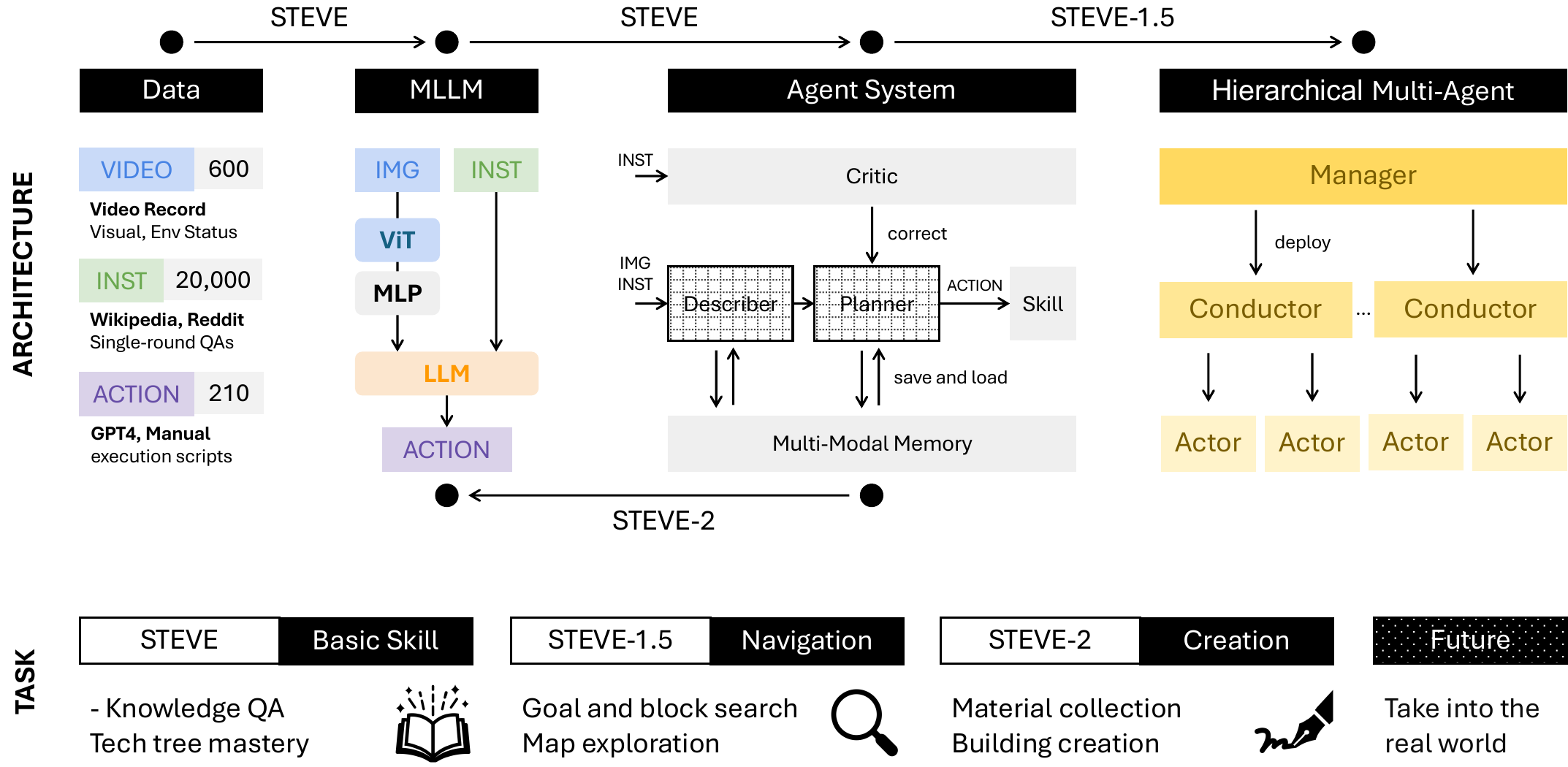}
    \captionof{figure}{\textbf{STEVE Series} overview.} 
    \label{fig:teaser}
\end{center}
}]


\begin{abstract}

Building an embodied agent system with a large language model (LLM) as its core is a promising direction. Due to the significant costs and uncontrollable factors associated with deploying and training such agents in the real world, we have decided to begin our exploration within the Minecraft environment. Our STEVE Series agents can complete basic tasks in a virtual environment and more challenging tasks such as navigation and even creative tasks, with an efficiency far exceeding previous state-of-the-art methods by a factor of $2.5\times$ to $7.3\times$. We begin our exploration with a vanilla large language model, augmenting it with a vision encoder and an action codebase trained on our collected high-quality dataset STEVE-21K. Subsequently, we enhanced it with a Critic and memory to transform it into a complex system. Finally, we constructed a hierarchical multi-agent system. Our recent work explored how to prune the agent system through knowledge distillation. In the future, we will explore more potential applications of STEVE agents in the real world. The code, data, and models are available at \href{https://rese1f.github.io/STEVE/}{site}.

\end{abstract}

\section{Introduction}


Recent advancements in artificial intelligence have seen the successful deployment of intelligent agents in the open-world game Minecraft, serving as a versatile platform for exploring complex agent behaviors and interactions~\citep{guss2019minerl,fan2022minedojo}. These agents, powered by large language models (LLMs)~\citep{chatgpt,touvron2023llama,song2023moviechat,song2024moviechat+}, demonstrate capabilities ranging from basic navigation to executing intricate tasks, embodying a significant leap in AI's potential for open-world understanding and interaction~\citep{yuan2023plan4mc,wang2023voyager，zhao2023survey}. Despite this progress, challenges remain in achieving seamless integration of multi-modal inputs and dynamic, autonomous decision-making that mirrors human-like intelligence and adaptability within such a variable and rich environment.



\section{Data and Environment}

The \textbf{STEVE-21K} dataset is integral for training the multi-modal Large Language Models (LLMs) in the \textbf{STEVE Series}, containing 600 Vision-Environment pairs, 20,000 Question-Answering pairs, and 210 Skill-Code pairs to enhance agents' interaction and task execution in Minecraft. Our simulation environment utilizes MineDojo~\citep{fan2022minedojo} and Mineflayer~\citep{mineflayer} APIs, providing a realistic setting for high-fidelity agent performance.

\section{Multi-Modal LLMs}

The \textbf{STEVE Series} advances through the integration of Multi-Modal Large Language Models (MLMs), essential for enhancing agent interactions within Minecraft. From \textbf{STEVE-1}~\citep{steve1}, using the fine-tuned STEVE-13B model, to \textbf{STEVE-2}~\citep{steve2} which incorporates advanced visual models like LLaVA~\citep{liu2023llava,liu2023improvedllava}, each version progressively enhances the agents' multimodal processing abilities.

\section{Hierarchical Multi-Agent System}

Introduced in \textbf{STEVE-1.5}, our Hierarchical Multi-Agent System enhances multi-agent cooperation for complex navigation and creation tasks in Minecraft. This system supports centralized planning and decentralized execution, enabling agents to adjust strategies and dynamically improve interaction with the environment. \textbf{STEVE-2} extends this system's capabilities, accommodating a broader range of activities and pushing the boundaries of autonomous multi-agent systems.

\section{Distill Embodied Agent into a Single Model}

\textbf{STEVE-2}~\citep{steve2} introduces a hierarchical knowledge distillation process that refines the alignment of tasks across various granularity levels within our agent system. This process incorporates the extra expert to enhance the teacher model with prior knowledge, significantly improving training quality for complex tasks. By distilling capabilities into a single model, \textbf{STEVE-2}~\citep{steve2} achieves operational simplicity and superior performance, setting a new benchmark in autonomous agent capabilities within Minecraft.

\section{Experiments}


\begin{table}[!t]
\centering
\resizebox{\linewidth}{!}{
\begin{tabular}{cc|cc}
\toprule
\multicolumn{2}{c|}{\bf Knowledge QA} & \multicolumn{2}{c}{\bf Tech Tree Mastery}\\
\cmidrule(lr){1-2} \cmidrule(lr){3-4}
 Model & preference~($\uparrow$) & Method & \# iters~($\downarrow$) \\
\midrule
Llama2-13B~\citep{touvron2023llama} & 6.89 & AutoGPT~\citep{autogpt} & 107 \\
GPT-4~\citep{openai2023gpt4} & 8.04 & Voyager~\citep{wang2023voyager} & 35 \\
\textbf{STEVE-13B}~\citep{steve1} & \textbf{8.12} & \textbf{STEVE-1}~\citep{steve1} & \textbf{33} \\
\bottomrule
\end{tabular}
}
\caption{\textbf{Comparison on Basic Skill}. Models preference rated 0-10 on knowledge QA and \# iters stand for average iterations for task fulfillment.}
\label{tab:navigation}
\end{table}

\subsection{Basic Skill}
The \textbf{STEVE series} demonstrates prowess in Knowledge Question and Answering and Tech Tree Mastery. STEVE-13B excels in producing precise Minecraft-related answers, surpassing both LLaMA2~\citep{touvron2023llama} and GPT-4~\citep{openai2023gpt4}. In Tech Tree Mastery, \textbf{STEVE-1}~\citep{steve1} progresses through Minecraft’s tech levels faster than competitors like AutoGPT~\citep{autogpt} and Voyager~\citep{wang2023voyager}, showcasing effective use of its vision unit to handle complex crafting tasks.

\subsection{Navigation}

\begin{table}[!t]
\centering
\resizebox{\linewidth}{!}{
\begin{tabular}{@{}l|c|c|c@{}}
\toprule
\multirow{2}{*}{Method} & \multirow{2}{*}{\# LLMs} & \bf Goal Search & \bf Map Explore\\
\cmidrule(lr){3-3} \cmidrule(lr){4-4}
& & success~($\uparrow$) & \# area~($\uparrow$)  \\
\midrule
\multirow{1}{*}{Voyager~\citep{wang2023voyager}}
                 & 12 / 20 & 64\% & 755 \\
\midrule
\multirow{1}{*}{STEVE-1~\citep{steve1}} 
              & 20 / 24  & 64\% & 696 \\ 
\midrule
\multirow{1}{*}{\textbf{STEVE-2}~\citep{steve2}}
                        & 5 / 8 & \textbf{91\%} & \textbf{1493}\\
\bottomrule
\end{tabular}
}
\caption{\textbf{Comparison on Navigation.} We list the success rate of Goal Search. \# area is the average squares of blocks over 5 iterations. We list the best performance with the number of LLMs for different tasks.}
\label{tab:navigation}
\end{table}

\textbf{STEVE-2}~\citep{steve2} excels in multi-modal goal search, continuous block search, and map exploration, outperforming existing models by substantial margins. In multi-modal goal search, STEVE-2 identifies goals using various sensory inputs with performance 5.5 $\times$ better than leading LLM-based methods. For map exploration, STEVE-2 updates and expands game maps with 1.9 $\times$ the efficiency of previous models, using a dynamic strategy tailored to unexplored areas.

\subsection{Creation}

\begin{table}[!t]
\centering
\resizebox{\linewidth}{!}{
\begin{tabular}{@{}l|c|c|c@{}}
\toprule
\multirow{2}{*}{Method} & \multirow{2}{*}{\# LLMs} & \bf Material Collection & \bf Building Creation\\
\cmidrule(lr){3-3} \cmidrule(lr){4-4}
& & completion~($\uparrow$) & FID~($\downarrow$)\\
\midrule
\multirow{1}{*}{Voyager}~\citep{wang2023voyager} & 4 & 72\% & 256.75 \\
\midrule
\multirow{1}{*}{Creative Agents~\citep{zhang2023creative}} & 4 & - & 68.32 \\
\midrule
\multirow{1}{*}{\textbf{STEVE-2}~\citep{steve2}} & 8 / 2 & \textbf{99\%} & \textbf{21.12} \\
\bottomrule
\end{tabular}
}
\caption{\textbf{Comparison on Creation.} We list task completion rates and average FID scores for image quality. We list the best performance with the number of LLMs for different tasks.}

\label{tab:creation}
\end{table}

In creation tasks, \textbf{STEVE-2}~\citep{steve2} significantly outperforms in material collection and building creation. It improves material gathering efficiency by 19 $\times$ over Voyager~\citep{wang2023voyager}. Additionally, using a finetuned VQ-VAE~\citep{razavi2019generating} for 3D occupancy generation, STEVE-2 enhances the quality of construction, achieving a 3.2 $\times$ increase in FID scores and surpassing other models and human evaluations in creative task performance.

\section{Conclusion}

The \textbf{STEVE series} has achieved substantial progress in multi-modal and hierarchical agent systems within Minecraft, excelling in tasks of basic skill, navigation, and creation.

\vspace{-10pt}

\paragraph{Future Work}
The next goal is to adapt the \textbf{STEVE series}' sophisticated agent technologies for practical applications in complex, dynamic real-world environments.

{\small
\bibliographystyle{ieee_fullname}
\bibliography{main}
}

\end{document}